\documentclass[10pt,twocolumn,letterpaper]{article}

\usepackage{iccv}
\usepackage{times}
\usepackage{epsfig}
\usepackage{graphicx}
\usepackage{amsmath}
\usepackage{amssymb}

\usepackage{multirow}
\usepackage{makecell}
\usepackage{tabulary}
\usepackage{booktabs}
\usepackage{bm}

\usepackage{tikz}
\usepackage{comment}
\usepackage{amsmath,amssymb} 
\usepackage{color}

\usepackage[colorlinks,bookmarks=false]{hyperref}
\usepackage{url}            
\usepackage{nicefrac}       
\usepackage{microtype}      
\usepackage{xcolor}         

\usepackage{subfigure}
\usepackage{wrapfig}
\usepackage{bbding}
\usepackage{colortbl} 

\newcommand{\tabincell}[2]{\begin{tabular}{@{}#1@{}}#2\end{tabular}}

\iccvfinalcopy 

\def\iccvPaperID{4189} 

\ificcvfinal\fi

\begin{document}

\title{SpectFormer: Frequency and Attention is what you need in a Vision Transformer}


\author{Badri Narayana Patro\\
Microsoft\\
{\tt\small badripatro@microsoft.com}
\and 
Vinay P. Namboodiri\\
University of Bath\\
{\tt\small vpn22@bath.ac.uk}
\and
Vijay Srinivas Agneeswaran\\
Microsoft\\
{\tt\small vagneeswaran@microsoft.com}
}

\maketitle
\ificcvfinal\thispagestyle{empty}\fi

\begin{abstract}
\vspace{-0.15in}
Vision transformers have been applied successfully for image recognition tasks. There have been either multi-headed self-attention based (ViT \cite{dosovitskiy2020image}, DeIT, \cite{touvron2021training}) similar to the original work in textual models or more recently based on spectral layers (Fnet\cite{lee2021fnet}, GFNet\cite{rao2021global}, AFNO\cite{guibas2021efficient}). We hypothesize that both spectral and multi-headed attention plays a major role. We investigate this hypothesis through this work and observe that indeed combining spectral and multi-headed attention layers provides a better transformer architecture. We thus propose the novel Spectformer architecture for transformers that combines spectral and multi-headed attention layers. We believe that the resulting representation allows the transformer to capture the feature representation appropriately and it yields improved performance over other transformer representations. For instance, it improves the top-1 accuracy by 2\% on ImageNet compared to both GFNet-H and LiT. SpectFormer-S reaches 84.25\% top-1 accuracy on ImageNet-1K (state of the art for small version). Further, Spectformer-L achieves 85.7\% that is the state of the art for the comparable base version of the transformers. We further ensure that we obtain reasonable results in other scenarios such as transfer learning on standard datasets such as CIFAR-10, CIFAR-100, Oxford-IIIT-flower, and Standford Car datasets.  We then investigate its use in downstream tasks such of object detection and instance segmentation on MS-COCO dataset and observe that Spectformer shows consistent performance that is comparable to the best backbones and can be further optimized and improved. Hence, we believe that combined spectral and attention layers are what are needed for vision transformers. The project page is available at this webpage.\url{https://badripatro.github.io/SpectFormers/}.

 
\end{abstract}

\vspace{-0.2in}
\section{Introduction}\label{intro}
 Transformers originated in natural language processing with the seminal work by Vaswani {\it et al.} \cite{vaswani2017attention}. Transformers have gone on to revolutionize the language domain in the form of large language models such as GPT-3 and its variants (including chatGPT)~\cite{gpt}, and Palm~\cite{chowdhery2022palm}. 
 Subsequent work extended the concept of Transformer to computer vision and other domains. 
 Interestingly, the main ideas explored retain much of the original transformer architecture. Clearly, the different domains could benefit from adapted transformers that are particular to the specific task. Through this work, we aim to specifically analyse the transformer for the  image classification task using a vision transformer. We show that the proposed adaptation, Spectformer, can outperform the state-of-the-art for this task.
\begin{figure}[tb]%
\centering
\includegraphics[width=0.49\textwidth]{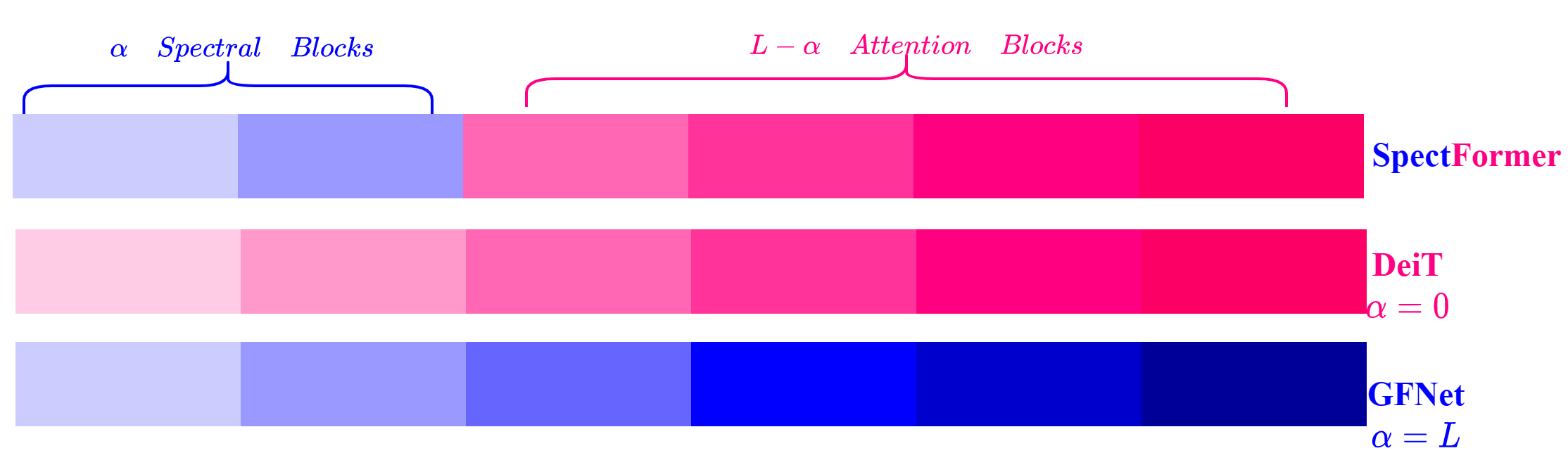}

\vspace{-0.4em}
\caption{SpectFormer architecture consist of L Blocks, out of which  $\alpha$ spectral blocks and  $L-\alpha$ attention blocks.}\label{fig_intro}
\vspace{-1.5em}
\end{figure}



The adaptation of transformers for  computer vision was first explored in the Vision Transformer (ViT) ~\cite{dosovitskiy2020image}. They made the important contribution of developing appropriate patch-based tokenization  for images whereby the transformer architecture could be used for images. DeIT~\cite{touvron2022deit} further improved the training process. The Fourier domain plays a major role in extracting frequency-based analysis of image information and has been well studied by the community. This is further supported by seminal work by Hubel and Weisel \cite{hubel1959receptive} that showed frequency tuned simple cells in the visual cortex. In transformers, it has been shown that the Fourier transforms could replace the multi-headed attention layers and achieve similar performance by Rao {\it et al.}~\cite{rao2021global} where they presented GFNet. They suggested that this approach captures fine-grained properties of images. This approach was further extended by AFNO~\cite{guibas2021efficient}, where they treated token mixing as operator learning.  We hypothesize that for the image domain, both spectral and multi-headed self-attention plays an important role. 

\begin{figure*}[htb]%
\centering
\includegraphics[width=0.84\textwidth]{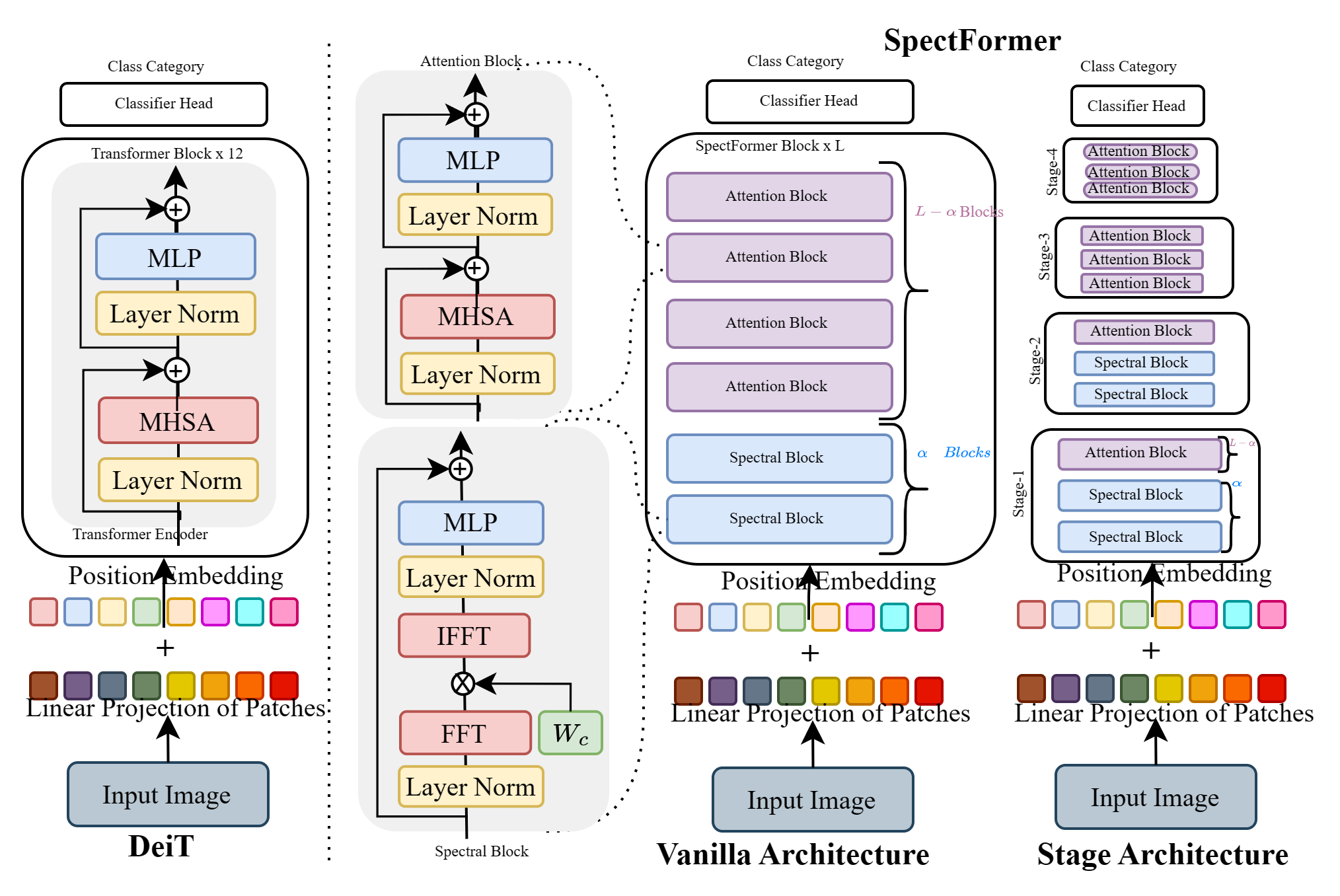}

\vspace{-0.4em}
\caption{This figure shows Architectural details of SpectFormer. The first part shows the DeiT\cite{touvron2021training} architecture. The second part shows the vanilla and Stage architecture of the SpectFormer Model. This also shows the layer structure of Spectral and Attention Blocks. }\label{fig_main}
\end{figure*}
There have also been a number of hierarchical transformer architectures that have been explored in the literature ~\cite{wang2021pyramid,chu2021twins, liu2021swin}.  
One of the hierarchical approaches has been LiT \cite{pan2022less} that uses less self-attention in the early layers, by using pure MLP (Multi-layer Perceptron) layers. They do use self-attention in deeper layers to capture longer dependencies. Motivated by the works related to spectral and also hierarchical transformers, we developed SpectFormer, a new transformer architecture that uses spectral layers implemented with Fourier Transform to capture relevant features in the initial layers of the architecture. Further, we use multi-headed self-attention in the deeper layers of the network. The SpectFormer architecture is simple and transforms the image tokens to the Fourier domain, then applies gating techniques using learnable weight parameters, and finally does the inverse Fourier transform to get the signal back. 
Our approach combines both spectral and multi-headed attention as shown in figure-\ref{fig_intro}.  

Our multi-headed self-attention layer is similar to the original attention paper \cite{vaswani2017attention}. We show that SpectFormer achieves state-of-art performance compared to parallel architectures like LiT and outperforms complete spectral architectures like GFNet~\cite{rao2021global} and AFNO\cite{guibas2021efficient}. It also outperforms complete multi-headed attention-based transformers like DeIT on ImageNet 1K dataset. We outline our contributions below:
\begin{itemize}

\item We design Spectformer by using initial spectral layers and multi-headed attention in deeper layers. We validate the choice of this architecture through thorough empirical validation. For instance, the visualisation of the learned filters for the spectral layers are more localised as compared to similar fully spectral GFT \cite{rao2021global}. The evidence suggests that adopting mixed spectral and later multi-headed attention results in improved results.   
We further validate this by comparing the proposed SpectFormer to a number of similar transformers such as LiT, vanilla transformers such as DeIT, spectral transformers such as GFNet, AFNO as well as hierarchical transformers such as PVT, Swin on the ImageNet dataset. 
\item We show that SpectFormer gets reasonable performance when used in transfer learning mode (trained on ImageNet and tested on CIFAR datasets) on CIFAR-10, and CIFAR-100 datasets.
\item Further, we show that SpectFormer obtains consistent performance in other tasks such as object detection and instance segmentation by evaluating its performance on the MS COCO dataset. 
\end{itemize}


\begin{table*}[!tb]
\scriptsize
\centering
\caption{Detailed architecture specifications for three variants of our SpectFormer with different model sizes, \emph{i.e.}, SpectFormer-S (small size), SpectFormer-B (base size), and SpectFormer-L (large size). $E_i$, $G_i$, $H_i$, and $C_i$ represent the expansion ratio of the feed-forward layer, the spectral gating number, the head number, and the channel dimension in each stage $i$, respectively.}
\setlength{\tabcolsep}{4.0pt}
\begin{tabular}{c|c|c|c|c}
\Xhline{2\arrayrulewidth}
        & O\/P Size & SpectFormer-H-S & SpectFormer-H-B & SpectFormer-H-L \\ \hline
Stage 1 & $\frac{H}{4} \times \frac{W}{4}$
        & $\left[ \begin{array}{c}  E_1=8 \\ G_1=1 \\ C_1=64  \end{array} \right] \!\times\! 2$ , $\left[ \begin{array}{c}  E_1=8 \\ H_1=2 \\ C_1=64  \end{array} \right] \!\times\! 1$
        & $\left[ \begin{array}{c}  E_1=8 \\ G_1=1 \\ C_1=64  \end{array} \right] \!\times\! 2$, $\left[ \begin{array}{c}  E_1=8 \\ H_1=2 \\ C_1=64  \end{array} \right] \!\times\! 1$
        & $\left[ \begin{array}{c}  E_1=8 \\ G_1=1 \\ C_1=96  \end{array} \right] \!\times\! 2$,$\left[ \begin{array}{c}  E_1=8 \\ H_1=3 \\ C_1=96  \end{array} \right] \!\times\! 1$
        \\ \hline
Stage 2 & $\frac{H}{8} \times \frac{W}{8}$
        & $\left[ \begin{array}{c} E_2=8 \\ G_2=1 \\  C_2=128 \end{array} \right] \!\times\! 2$ ,$\left[ \begin{array}{c} E_2=8 \\ H_2=4 \\  C_2=128 \end{array} \right] \!\times\! 2$
        & $\left[ \begin{array}{c} E_2=8 \\ G_2=1 \\  C_2=128 \end{array} \right] \!\times\! 2$, $\left[ \begin{array}{c} E_2=8 \\ H_2=4 \\  C_2=128 \end{array} \right] \!\times\! 2$
        & $\left[ \begin{array}{c} E_2=8 \\ G_2=1 \\  C_2=192 \end{array} \right] \!\times\! 2$, $\left[ \begin{array}{c} E_2=8 \\ H_2=6 \\  C_2=192 \end{array} \right] \!\times\! 4$
        \\ \hline
Stage 3 & $\frac{H}{16} \times \frac{W}{16}$
        & $\left[ \begin{array}{c}  E_3=4 \\ H_3=10 \\ C_3=320 \end{array} \right] \!\times\! 6$
        & $\left[ \begin{array}{c}  E_3=4 \\ H_3=10 \\ C_3=320 \end{array} \right] \!\times\! 12$
        & $\left[ \begin{array}{c}  E_3=4 \\ H_3=12 \\ C_3=384 \end{array} \right] \!\times\! 18$
        \\ \hline
Stage 4 & $\frac{H}{32} \times \frac{W}{32}$
        & $\left[ \begin{array}{c} E_4=4 \\ H_4=14 \\ C_4=448 \end{array} \right] \!\times\! 3$
        & $\left[ \begin{array}{c} E_4=4 \\ H_4=16 \\ C_4=512 \end{array} \right] \!\times\! 3$
        & $\left[ \begin{array}{c} E_4=4 \\ H_4=16 \\ C_4=512 \end{array} \right] \!\times\! 3$
        \\ \Xhline{2\arrayrulewidth}
\end{tabular}
\label{tab:architecture}
\end{table*}

\section{Related Work}\label{related}

\textbf{Quadratic Complexity of Attention Nets:} The Vision Transformer (ViT)~\cite{dosovitskiy2020image} model considers the image as a 16x16 word and is used to classify the image into predefined categories. In the ViT model, each image is split into a sequence of tokens of fixed length and then applied to multiple transformer layers to capture the global relationship across the token for the classification task. Touvron et al. ~\cite{touvron2021training} proposed an efficient transformer model based on distillation technique (DeiT). It uses a teacher-student strategy that relies on a distillation token to ensure that a student learns from a teacher through attention. Bao et al.~\cite{bao2021beit} have proposed a masked image model task for a pretrained vision transformer. The vanilla transformer architecture which uses multi-headed self-attention (MSA) for efficient token mixing includes papers such as Tokens-to-token ViT \cite{yuan2021tokens}, Transformer iN Transformer (TNT)~\cite{han2021transformer},  Cross-ViT~\cite{chen2021crossViT}, Class attention image Transformer(CaiT) \cite{touvron2021going} etc. The architectural complexity of most of the above transformers is O(N$^2$).  Attempts at alleviating this include the Uniformer \cite{li2022uniformer} which brings the best of convolutional nets and transformers by using multi-headed relation aggregation and RegionViT \cite{chen2022regionvit} as well as Token Pyramid Vision Transformer (TopFormer) \cite{zhang2022topformer}. A recent effort to address this complexity was by the paper \cite{choukroun2022error} which used scaled element-wise embedding and an adapted-mask self-attention and enabled transformers to solve error-correcting codes. The complexity has also been mitigated by using spectral transformers, which typically have O($N$log$N$) complexity. They also reduce the parameter count significantly compared to vanilla transformers.

\begin{figure}[tb]%
\centering
\includegraphics[width=0.4243\textwidth]{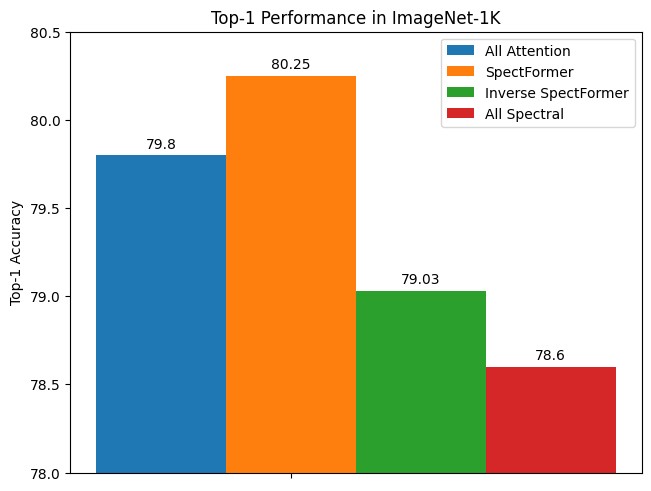}
\vspace{-0.4em}
\caption{Initial comparison of attention vs spectral combinations.}
\label{fig:ana}
\vspace{-1.5em}
\end{figure}

\begin{figure*}[]
    \begin{minipage}[t]{.492345\textwidth}
        \centering
        \includegraphics[width=0.4943\textwidth]{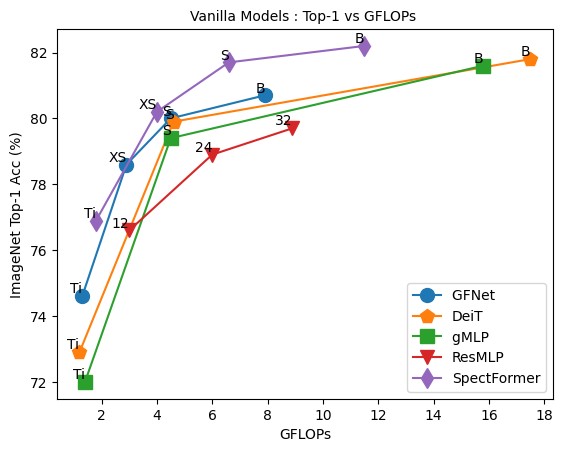}
            \includegraphics[width=0.4943\textwidth]{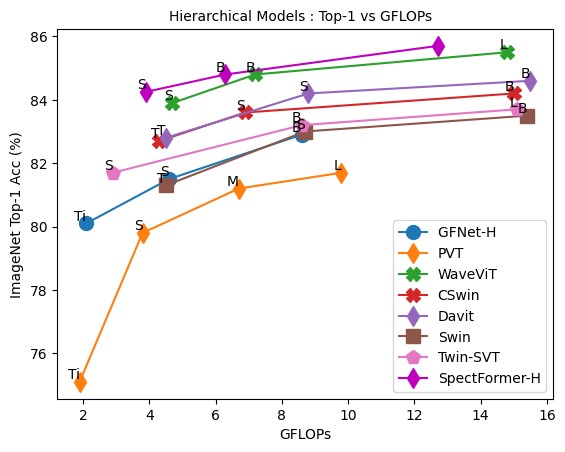}
            \caption{Comparison of ImageNet Top-1 Accuracy (\%) vs GFLOPs of various models in Vanilla and Hierarchical architecture.}
            \label{fig:gflops}
    \end{minipage}
    \hfill
    \begin{minipage}[t]{.492345\textwidth}
        \centering
        \includegraphics[width=0.4943\textwidth]{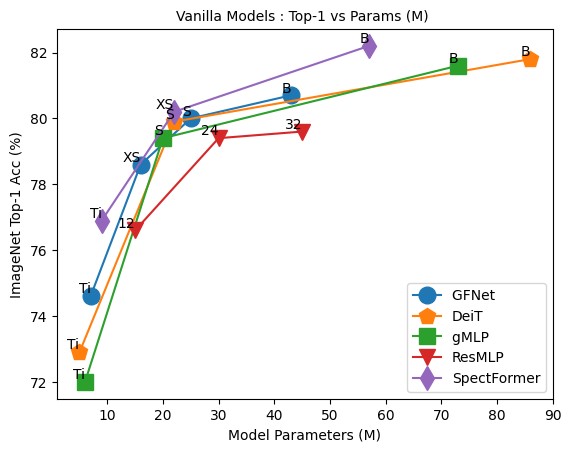}
            \includegraphics[width=0.4943\textwidth]{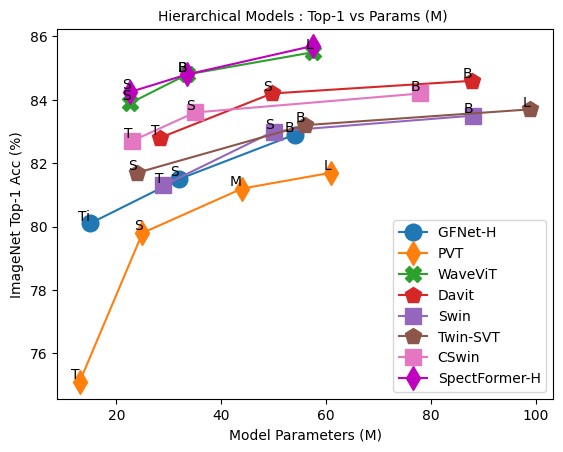}
            \caption{Comparison of ImageNet Top-1 Accuracy (\%) vs Parameters (M) of various models in Vanilla and Hierarchical architecture.}
            \label{fig:param}
    \end{minipage}  
    \label{fig:1-2}
\end{figure*}


\textbf{Spectral Transformers}
 Inspired by an MLP-mixer-based token mixing technique, recent  work carried out using a spectral mixing technique in which  the self-attention layer of the transformer is replaced by a non-parameterized Fourier transformation (Fnet)~\cite{lee2021fnet}, which is then followed by a non-linearity and feed-forward network. This was followed by the Global Filter network (GFNet)~\cite{rao2021global}, which uses a depth-wise global convolution for token mixing.  Guibias et al. ~\cite{guibas2021efficient} formulated the token mixing task as an operator-learning task that learns mapping among continuous functions in infinite dimensional space using Fourier Neural Operator (FNO) ~\cite{li2020fourier}.  In Wave-ViT~\cite{yao2022wave}, the author has discussed the quadratic complexity of the self-attention network of the transformer model with input patch numbers. They have proposed a wavelet vision transformer to perform lossless down-sampling using wavelet transform over keys and values of the self-attention network. The model obtains state-of-the-art results on image recognition, object detection, and instance segmentation tasks. Recently, another work \cite{nguyen2022fourierformer}
proposes a Fourier integral theorem characterize attention as non-parametric kernel regression and approximate key-query distributions. We compare the performance of the above methods with the proposed SpectFormer model.

\textbf{Hierarchical Transformers}
Hierarchical architecture-based transformers are widely used to improve the performance of the transformer. It is a four-stage architecture first proposed by Wang et al. ~\cite{wang2021pyramid} in Pyramid vision Transformer (PVT). Later the stage-based architecture is used by SwinT ~\cite{liu2021swin}, Twins ~\cite{chu2021twins} transformers.
Through the work on LiT \cite{pan2022less}, Pan {\it et al.} have proposed an attention-based hierarchical transformer architecture that used to pay less attention in vision transformers using MLP layers. It must be noted that SpectFormer, which uses Fourier-based spectral layers is a more generalized and efficient mechanism to capture localized features that are fine-grained components of the image. The initial spectral layers improve the performance and achieve state-of-art compared to WaveViT \cite{yao2022wave} as mentioned in the recent efficient360 work\cite{patro2023efficiency}.  We compare the performance of SpectFormer with recent works and  observe that the proposed work outperforms all these other works.

\begin{table*}[]
  \centering
\caption{In this table, we present detailed configurations of various versions of SpectFormer for the vanilla transformer architecture. The table provides information on the number of heads, embedding dimensions, the number of layers in each variant, and the training resolution. For hierarchical SpectFormer-H models, we provide information in Table-1 of the main paper, which includes details for four stages. The FLOPs (floating-point operations) are calculated for both $224\times 224$ and $384\times 384$ input sizes. For the vanilla SpectFormer architecture, we use four spectral layers with $\alpha=4$, while the remaining attention layers are equal to $(L-\alpha)$.} \vspace{5pt}
    \begin{tabular}{c|c|c|c|c|c|c}
    \toprule
    Model & \#Layers &  \#heads &  \#Embedding Dim & Params (M) & Training Resolution & FLOPs (G)  \\ \midrule
    SpectFormer-Ti &  12 & 4& 256  & 9 & 224& 1.8\\
    SpectFormer-XS &  12& 6 & 384  & 20& 224 & 4.0\\
    SpectFormer-S &  19 & 6& 384  & 32& 224 & 6.6\\
    SpectFormer-B &  19 & 8& 512  & 57 & 224& 11.5\\
      \midrule
    SpectFormer-XS &  12& 6 & 384  & 21& 384 & 13.1\\
    SpectFormer-S &  19 & 6& 384  & 33& 384 & 22.0\\
    SpectFormer-B &  19 & 8& 512  & 57& 384& 37.3\\
      \bottomrule
    \end{tabular}%
  \label{tab:arch} 
\end{table*}%

\section{Method: Spectformer}
\subsection{Need for mixed spectral transformer}

In order to validate whether there is a difference in performance in terms of representation if we consider a mixture of spectral and multi-headed attention layers, we did a study by considering the difference between basic all-attention, all-spectral and mixed spectral-attention layers with the spectral layers being placed initially or finally (that we term inverse SpectFormer). This analysis is provided in the following figure~\ref{fig:ana}. As can be observed, the particular configuration where we use spectral layers initially followed by multi-headed attention layers is more beneficial. Thus, the need for Spectformer is evident. We therefore propose an architecture that includes initial spectral layers followed by multi-headed attention layers.

\subsection{SpectFormer Architecture}
The key idea of the SpectFormer architecture is illustrated in figure \ref{fig_main}. As can be inferred from the figure, the SpectFormer architecture comprises a patch embedding layer, followed by a positional embedding layer, followed by a transformer block, and followed by a classification head (MLP - dim to 1000 projection). The transformer block comprises a series of spectral layers followed by attention layers. The image is split into a sequence of patches and we obtain a patch embedding using a linear projection layer. Our positional embedding uses a standard positional encoding layer. The transformer block will be explained below in two parts.

\subsection{Spectral Block}
The objective of the spectral layer is to capture the different frequency components of the image to comprehend localized frequencies. This can be achieved using a spectral gating network, that comprises a Fast Fourier Transform (FFT) layer, followed by a weighted gating, followed by an inverse FFT layer. The spectral layer converts physical space into the spectral space using FFT. We use a learnable weight parameter to determine the weight of each frequency component so as to capture the lines and edges of an image appropriately. The learnable weight parameter is specific to each layer of SpectFormer and is learnt using back-propagation techniques.

The spectral layer uses an inverse Fast Fourier Transform (IFFT) to bring back the spectral space back to the physical space. Following the IFFT, the spectral layer has layer normalization and Multi-Layer Perceptron (MLP) block for channel mixing, while token mixing is done using the spectral gating technique. Note that while we use the FFT/IFFT, the method can also be implemented using wavelet/inverse wavelet transform.

\subsection{Attention Block}

The attention layer of SpectFormer is a standard attention layer comprising layer normalization, followed by multi-headed self-attention (MHSA), followed by layer normalization and is followed by an MLP. The MHSA architecture is similar to DeIT attention architecture in that
MHSA is used for token mixing and MLP is used for channel mixing in the attention layer.

\subsection {SpectFormer Block}

SpectFormer block has been illustrated in the figure \ref{fig_intro}, in the staged architecture. We introduce an alpha factor in the SpectFormer block, which controls the number of spectral layers and attention layers. If $\alpha$=0, SpectFormer comprises all attention layers, similar to DeIT-s, while with an $\alpha$ value of 12, SpectFormer becomes similar to GFNet, with all spectral layers. It must be noted that all attention layers have the disadvantage that local features cannot be captured accurately. 
Similarly, all spectral layers have the disadvantage that global image properties or semantic features cannot be handled accurately. SpectFormer gives the flexibility to vary the number of spectral and attention layers, which helps in capturing both global properties as well as local features accurately. Consequently, as also ratified by our performance studies, SpectFormer considers local features, which help to capture localized frequencies in initial layers as well as global features in the deeper layers, which help capture long-range dependencies. 

The above explanation is mainly for the vanilla SpectFormer architecture. We have also come up with a staged architecture, which comprises four stages, with each stage having a varying number of SpecfFormer blocks. Stage 1 has 3 SpectFormer blocks, while Stage 2 has 4, Stage 3 has 6 and Stage 4 has 3 SpectFormer blocks, as shown in table \ref{tab:architecture}. In the stage of SpectFormer-s, there are 2 spectral layers and 1 attention layer, while Stage 2 comprises 2 spectral and 2 attention layers, to capture the local information. Stages 3 and 4 comprise only attention layers, to capture the semantic information. The details of SpectFormer-s architecture were explained above, while SpectFormer-B and SpectFormer-L are depicted in the table. We came up with several variants of the spectral layer including using FNet, FNO, GFNet and AFNO.  We also provide a details  SpectFormer  architecture of  the vanilla transformer model, presented in table-\ref{tab:arch}.


\section{Experiments and Results}
 Our proposed SpectFormer, is evaluated through various empirical evidence on a range of mainstream computer vision tasks, including image recognition, object detection, and instance segmentation. To compare the quality of learned feature representations obtained from SpectFormer, we conduct the following evaluations: (a) Conducting ablation studies that support each variant in our SpectFormer  block and select best $\alpha$ value  for it; (b) Training from scratch for image recognition task on ImageNet1K; (c) Transfer learning on CIFAR10, CIFAR-100, Oxford-IIIT flower, Standford Car dataset for Image recognition task using the SpectFormer (pre-trained on ImageNet1K) model; (d) Fine-tuning the SpectFormer (pre-trained on ImageNet1K) for downstream tasks such as object detection and instance segmentation on COCO;  and (e) Visualizing the learned visual representation by SpectFormer. Through these evaluations, we demonstrate the effectiveness of our proposed SpectFormer model for computer vision tasks.

\begin{table}[htb]
\centering
\caption{This table shows the ablation analysis of various spectral layers in SpectFormer architecture such as the Fourier Network (FN), the Fourier Gating Network (FGN), the Wavelet Gating Network (WGN), and the Fourier Neural Operator (FNO). We conduct this ablation study on the small-size networks in stage architecture. This indicates that FGN performs better than other kinds of networks.}\label{Exp_tab_1}
\begin{tabular}{lcccc}
\hline
Model &  \tabincell{c}{Params \\ (M)} &  \tabincell{c}{FLOPs \\ (G)}  &  \tabincell{c}{Top-1 \\ (\%)}   &  \tabincell{c}{Top-5 \\ (\%)} \\
\hline
SpectFormer\_FN & 21.17 &3.9& 84.02& 96.77\\
SpectFormer\_FNO & 21.33 &3.9& 84.09&96.86\\
SpectFormer\_WGN & 21.59& 3.9& 83.70&96.56\\
SpectFormer\_FGN & 22.22& 3.9& \textbf{84.25}&\textbf{96.93}\\
\hline
\end{tabular}

\end{table}

\begin{table}[]
\centering
\caption{Ablation Analysis with different variants of SpectFormer architecture with variying alpha.}\label{Exp_tab_2}
\begin{tabular}{lcccc}
\hline
Model &  \tabincell{c}{Params \\ (M)} &  \tabincell{c}{FLOPs \\ (G)}  &  \tabincell{c}{Top-1 \\ (\%)}   &  \tabincell{c}{Top-5 \\ (\%)} \\
\hline
SpectFormer\_$\alpha_{0}$ & 22.00 &4.6& 79.80& -\\
SpectFormer\_$\alpha_{2}$ &21.03 & 4.3& 79.87& 94.69\\
SpectFormer\_$\alpha_{4}$ & 20.02 &4.0& \textbf{80.21}& 94.76\\
SpectFormer\_$\alpha_{6}$ & 19.01 &3.7& 80.14& \textbf{94.85}\\
SpectFormer\_$\alpha_{8}$ & 18.00 &3.4& 79.55.& 94.59\\
SpectFormer\_$\alpha_{10}$ & 16.99 &3.1& 79.06& 94.62\\
SpectFormer\_$\alpha_{12}$ & 16.00 &2.9& 78.60& 94.20\\\hline
iSpectFormer\_$\alpha_{4}$ & 20.02 &4.0& 79.03& 94.30\\
\hline
\end{tabular}

\end{table}

\begin{table}[]
\caption{This table shows the SpectFormer performance based on different model size. The first part shows results on vanilla architecture for Fourier Gating Network based model. The second part shows results for Hierarchical architecture indicated by 'H'. These results are for $\alpha=4$.}\label{tab:variants}
\centering
\setlength{\tabcolsep}{4.0pt}
\begin{tabular}{lcccc}

\hline
Model &  \tabincell{c}{Params \\ (M)} &  \tabincell{c}{FLOPs \\ (G)}  &  \tabincell{c}{Top-1 \\ (\%)}   &  \tabincell{c}{Top-5 \\ (\%)} \\
\hline

SpectFormer-T & 9.15& 1.8& 76.89 &93.38\\
SpectFormer-XS & 20.02& 4.0& 80.21 &94.76\\
SpectFormer-S & 32.56& 6.6& 81.70 &95.64\\
SpectFormer-B &57.15& 11.5& \textbf{82.12}& \textbf{95.75}\\ \hline

SpectFormer-H-FN-S & 21.17& 3.9 & 84.02 &96.77\\
SpectFormer-H-FN-B &31.99&6.3& 85.04& 97.37\\  
SpectFormer-H-WGN-S & 22.59&  3.9& 83.7 &96.56\\
SpectFormer-H-WGN-B &33.42& 6.3& 84.57& 96.97\\ 
SpectFormer-H-S & 22.22& 3.9& 84.25 &96.93\\
SpectFormer-H-B &33.05& 6.3& 85.05&97.30\\
SpectFormer-H-L &54.67&12.7 &\textbf{85.7}&\textbf{ 97.52}\\\hline

\hline
\end{tabular}
\vspace{-0.2in}
\end{table}

\begin{table}[t]
\caption{This shows a performance comparison of SpectForm with similar Transformer Architecture with different sizes of the networks on ImageNet-1K. $\star$ indicates additionally trained with the Token Labeling objective using MixToken\cite{jiang2021all}.}
\centering
\setlength{\tabcolsep}{4.0pt}
\begin{tabular}{l|c|c|c|cc}
    \toprule
    Network  &  Params   &GFLOPs  &  Top-1    &  Top-5   \tabularnewline
    \midrule
    \multicolumn{5}{c}{Vanilla Transformer Comparison}  \\ 
    \midrule		
    
%
 
DeiT-Ti ~\cite{touvron2021training}      & 5M & 1.2&  72.2&  91.1\\
FourierFormer~\cite{nguyen2022fourierformer} & - & -&  73.3&  91.7\\
GFNet-Ti~\cite{rao2021global}      & 7M&  1.3&  74.6 & 92.2\\
\rowcolor{gray!15}SpectFormer-T & 9M& 1.8& \textbf{76.9} &\textbf{93.4}\\ \hline

DeiT-S ~\cite{touvron2021training}      & 22M&  4.6&  79.8 & 95.0\\
Fnet-S ~\cite{lee2021fnet}   & 15M &2.9&  71.2 &-\\
GFNet-XS~\cite{rao2021global}      & 16M&  2.9 &  78.6&  94.2\\ 
GFNet-S ~\cite{rao2021global}      &25M&  4.5&  80.0 & 94.9\\
\rowcolor{gray!15}SpectFormer-XS & 20M& 4.0&\textbf{ 80.2} &\textbf{94.7}\\
\rowcolor{gray!15}SpectFormer-S & 32M& 6.6&\textbf{ 81.7} &\textbf{95.6}\\
\hline

DeiT-B ~\cite{touvron2021training}   & 86M& 17.5 & 81.8& 95.6\\ 
GFNet-B ~\cite{rao2021global}   & 43M &7.9& 80.7 &95.1\\
\rowcolor{gray!15}SpectFormer-B &57M& 11.5& \textbf{82.1}& \textbf{95.7}\\
\midrule
\multicolumn{5}{c}{Hierarchical Transformer Comparison}  \tabularnewline
\midrule

PVT-S~\cite{wang2021pyramid} & 25M & 3.8 &79.8 &-\\
Swin-T~\cite{liu2021swin}   &  29M & 4.5&  81.3    & - \tabularnewline
GFNet-H-S~\cite{rao2021global} &32M& 4.6& 81.5& 95.6\\
LIT-S~\cite{pan2022less} & 27M&  4.1&  81.5 & &-\\
\rowcolor{gray!15}{SpectFormer-H-S} & 21.7M & 3.9  & 83.1 & 96.3 &\\
\rowcolor{gray!15}SpectFormer-H-S$^\star$ & 22M & 3.9 & \textbf{84.2} &\textbf{96.9}\\ \hline

PVT-M~\cite{wang2021pyramid} & 44M& 6.7& 81.2& -\\
Swin-S~\cite{liu2021swin}    &  50M & 8.7&  83.0    & - \tabularnewline
GFNet-H-B~\cite{rao2021global} &54M& 8.6 &82.9& 96.2\\
LIT-M~\cite{pan2022less} & 48M&  8.6&  83.0& &-\\
\rowcolor{gray!15}SpectFormer-H-B$^\star$ & 33M & 6.3 & \textbf{85.0}&\textbf{97.3}\\ \hline

PVT-L~\cite{wang2021pyramid}   &  61M & 9.8 &  82.3  & -  \tabularnewline
Swin-B~\cite{liu2021swin}      &  88M & 15.4&  83.3   & - \tabularnewline
LIT-B~\cite{pan2022less} & 86M & 15.0 & 83.4 & -\\
\rowcolor{gray!15} SpectFormer-H-L$^\star$ &55M & 12.7 &\textbf{85.7}& \textbf{97.5}\\\hline



\end{tabular}
\label{tab:similar_arch_imagenet}
\end{table}

\begin{table*}[!tb]
\centering
\caption{The table shows the performance of various vision backbones on the ImageNet1K\cite{deng2009imagenet} dataset for image recognition tasks.  $\star$ indicates additionally trained with the Token Labeling objective using MixToken\cite{jiang2021all} and a convolutional stem (conv-stem) \cite{wang2022scaled} for patch encoding. We have grouped the vision models into three categories based on their GFLOPs (Small, Base, and Large). The GFLOP ranges:Small (GFLOPs$<$6), Base (6$\leq$GFLOPs$<$10), Large (10$\leq$GFLOPs$<$30).
}
\setlength{\tabcolsep}{4.0pt}
\begin{tabular}{c|c|c|cc|c|cc|cc}
\Xhline{2\arrayrulewidth}
Method          & Params & GFLOPs & Top-1 & Top-5 & Method          & Params & GFLOPs & Top-1 & Top-5 \\ \hline
\multicolumn{5}{c|} {Small} & \multicolumn{5}{c} {Large} \\ \hline
ResNet-50  \cite{he2016deep}                 & 25.5M & 4.1 & 78.3 & 94.3 &
ResNet-152 \cite{he2016deep}                 & 60.2M & 11.6 & 81.3 & 95.5  \\

BoTNet-S1-50 \cite{srinivas2021bottleneck}   & 20.8M & 4.3 & 80.4 & 95.0 &
ResNeXt101 \cite{xie2017aggregated}    & 83.5M & 15.6 & 81.5 & -     \\
Cross-ViT-S~\cite{chen2021crossViT}& 26.7M& 5.6 & 81.0 &- &
gMLP-B~\cite{liu2021pay}                & 73.0M& 15.8&81.6&-\\

Swin-T  \cite{liu2021swin}                   & 29.0M & 4.5 & 81.2 & 95.5 & 
DeiT-B \cite{touvron2021training}            & 86.6M & 17.6 & 81.8 & 95.6  \\

ConViT-S \cite{d2021convit}                  & 27.8M & 5.4 & 81.3 & 95.7 &
SE-ResNet-152 \cite{hu2018squeeze}           & 66.8M & 11.6 & 82.2 & 95.9  \\

T2T-ViT-14 \cite{yuan2021tokens}             & 21.5M & 4.8 & 81.5 & 95.7 &
Cross-ViT-B~\cite{chen2021crossViT} & 104.7M& 21.2 & 82.2  &- \\

RegionViT-Ti+ \cite{chen2022regionvit}       & 14.3M & 2.7 & 81.5 & -    & 
ResNeSt-101 \cite{zhang2022resnest}          & 48.3M & 10.2 & 82.3 & -     \\

SE-CoTNetD-50  \cite{li2022contextual}       & 23.1M & 4.1 & 81.6 & 95.8    &
ConViT-B \cite{d2021convit}                  & 86.5M & 16.8 & 82.4 & 95.9  \\

Twins-SVT-S \cite{chu2021twins}              & 24.1M & 2.9 & 81.7 & 95.6 &
PoolFormer-M48 ~\cite{yu2022metaformer} & 73.0M&  11.8&  82.5&-\\

CoaT-Lite Small \cite{xu2021co}              & 20.0M & 4.0 & 81.9 & 95.5 &
T2T-ViTt-24 \cite{yuan2021tokens}            & 64.1M & 15.0 & 82.6 & 95.9  \\

PVTv2-B2 \cite{wang2022pvt}                & 25.4M & 4.0 & 82.0 & 96.0 & 

TNT-B  \cite{han2021transformer}             & 65.6M & 14.1 & 82.9 & 96.3  \\

LITv2-S~\cite{panfast} &28.0M &3.7& 82.0&-  & 
CycleMLP-B4~\cite{chencyclemlp} &52.0M& 10.1& 83.0&- 
\\

MViTv2-T~\cite{li2022mvitv2}&24.0M& 4.7& 82.3&-  &
DeepViT-L  \cite{zhou2021deepvit}            & 58.9M & 12.8 & 83.1 & -     \\

Wave-ViT-S~\cite{yao2022wave}                & 19.8M & 4.3 & 82.7 & 96.2 & 
RegionViT-B \cite{chen2022regionvit}         & 72.7M & 13.0 & 83.2 & 96.1  \\

CSwin-T \cite{dong2022cswin}             & 23.0M & 4.3 & 82.7 & - &
CycleMLP-B5~\cite{chencyclemlp}             & 76.0M& 12.3& 83.2&-  \\

DaViT-Ti  ~\cite{ding2022daViT}         & 28.3M & 4.5 &  82.8&- &
ViP-Large/7 ~\cite{hou2022vision}       & 88.0M& 24.4& 83.2&- \\

\cellcolor{gray!20}{SpectFormer-H-S} & \cellcolor{gray!20}{21.7M} & \cellcolor{gray!20}{3.9}  & \cellcolor{gray!20}{83.1} & \cellcolor{gray!20}{96.3} &
CaiT-S36 \cite{touvron2021going}             & 68.4M  & 13.9 & 83.3 & -     \\

 iFormer-S\cite{si2022inception} & 20.0M & 4.8 & 83.4 & 96.6 &
AS-MLP-B ~\cite{lianmlp} & 88.0M& 15.2& 83.3&-\\

CMT-S~\cite{guo2022cmt}& 25.1M &  4.0& 83.5 &- &
BoTNet-S1-128 \cite{srinivas2021bottleneck}  & 75.1M & 19.3 & 83.5 & 96.5  \\ 

 MaxViT-T~\cite{tu2022maxvit}           & 31.0M&  5.6&  83.6&- &
 Swin-B  \cite{liu2021swin}                   & 88.0M & 15.4 & 83.5 & 96.5  \\

 Wave-ViT-S$^\star$~\cite{yao2022wave}       & 22.7M & 4.7 & 83.9 & 96.6 & 
 Wave-MLP-B~\cite{tang2022image}& 63.0M& 10.2& 83.6&-  \\

\cellcolor{gray!15}\textbf{SpectFormer-H-S$^\star$} & \cellcolor{gray!15}\textbf{22.2M} & \cellcolor{gray!15}\textbf{3.9}  & \cellcolor{gray!15}\textbf{84.3} & \cellcolor{gray!15}\textbf{96.9} &
LITv2-B ~\cite{panfast} & 87.0M &13.2 & 83.6&- \\
 \cline{1-5}
\multicolumn{5}{c|} {Base}& 
 PVTv2-B4 \cite{wang2022pvt}                & 62.6M & 10.1 & 83.6 & 96.7  \\
\cline{1-5}

ResNet-101 \cite{he2016deep}                 & 44.6M & 7.9 & 80.0 & 95.0 &
ViL-Base ~\cite{zhang2021multi}             & 55.7M&  13.4 &83.7& -\\

BoTNet-S1-59 \cite{srinivas2021bottleneck}   & 33.5M & 7.3 & 81.7 & 95.8 &
Twins-SVT-L \cite{chu2021twins}              & 99.3M & 15.1 & 83.7 & 96.5  \\

T2T-ViT-19 \cite{yuan2021tokens}             & 39.2M & 8.5 & 81.9 & 95.7 &
Hire-MLP-Large~\cite{Guo_2022_CVPR} & 96.0M& 13.4& 83.8&-  \\

CvT-21~\cite{wu2021CvT}                     & 32.0M & 7.1 & 82.5 & -    & 
RegionViT-B+ \cite{chen2022regionvit}        & 73.8M & 13.6 & 83.8 & -     \\

GFNet-H-B~\cite{rao2021global}             &54.0M& 8.6 &82.9& 96.2&
Focal-Base \cite{yang2021focal}              & 89.8M & 16.0 & 83.8 & 96.5  \\

Swin-S  \cite{liu2021swin}                   & 50.0M & 8.7 & 83.2 & 96.2 &
PVTv2-B5 \cite{wang2022pvt}                & 82.0M & 11.8 & 83.8 & 96.6  \\

Twins-SVT-B \cite{chu2021twins}              & 56.1M & 8.6 & 83.2 & 96.3 &
SE-CoTNetD-152 \cite{li2022contextual} & 55.8M & 17.0 & 84.0 & 97.0 \\

SE-CoTNetD-101 \cite{li2022contextual}     & 40.9M & 8.5   & 83.2 & 96.5 &
 DAT-B  ~\cite{xia2022vision} &  88.0M& 15.8&  84.0&-\\

PVTv2-B3 \cite{wang2022pvt}                & 45.2M & 6.9 & 83.2 & 96.5 &
LV-ViT-M$^\star$ \cite{jiang2021all}         & 55.8M & 16.0 & 84.1 & 96.7  \\

LITv2-M~\cite{panfast} & 49.0M& 7.5 & 83.3&-  &
 CSwin-B ~\cite{dong2022cswin} & 78.0M  &15.0  & 84.2&- \\ 

RegionViT-M+ \cite{chen2022regionvit}        & 42.0M & 7.9 & 83.4 & -    & 
 HorNet-$B_{GF}$~\cite{rao2022hornet}& 88.0M& 15.5& 84.3&-  \\

MViTv2-S~\cite{li2022mvitv2} &35.0M & 7.0 & 83.6&- &
DynaMixer-L~\cite{wang2022dynamixer}        & 97.0M& 27.4& 84.3&- \\

 CSwin-S ~\cite{dong2022cswin}& 35.0M & 6.9  & 83.6&- &
MViTv2-B~\cite{li2022mvitv2} &52.0M& 10.2& 84.4&- \\

DaViT-S ~\cite{ding2022daViT}& 49.7M & 8.8 & 84.2&- &
DaViT-B~\cite{ding2022daViT}& 87.9M & 15.5  &84.6&- \\

VOLO-D1$^\star$  \cite{yuan2022volo}         & 26.6M & 6.8 & 84.2 & - &
CMT-L ~\cite{guo2022cmt}&74.7M&19.5& 84.8&-\\

CMT-B ~\cite{guo2022cmt}&45.7M&9.3& 84.5&-&
MaxViT-B ~\cite{tu2022maxvit}& 120.0M& 23.4&85.0&- \\

MaxViT-S~\cite{tu2022maxvit}& 69.0M& 11.7& 84.5&- &
VOLO-D2$^\star$ \cite{yuan2022volo}          & 58.7M & 14.1 & 85.2 & -     \\

iFormer-B\cite{si2022inception} & 48.0M & 9.4 & 84.6 & 97.0 &
VOLO-D3$^\star$ \cite{yuan2022volo}          & 86.3M & 20.6 & 85.4 & -     \\

Wave-ViT-B$^\star$  \cite{yao2022wave}& 33.5M & 7.2 & 84.8 & 97.1 &
Wave-ViT-L$^\star$ \cite{yao2022wave}  & 57.5M & 14.8 & 85.5 & 97.3  \\

\rowcolor{gray!15}\textbf{SpectFormer-H-B$^\star$} & \textbf{33.1M} & \textbf{6.3}& \textbf{85.1} & \textbf{97.3} & \textbf{SpectFormer-H-L$^\star$} & \textbf{54.7M} & \textbf{12.7} & \textbf{85.7} & \textbf{97.5} \\
\Xhline{2\arrayrulewidth}
\end{tabular}
\label{tab:imagenet1k_sota}
\vspace{-0.23in}
\end{table*}






\subsection{Ablation analysis on spectral architectures}
We conduct an experiment on the spectral network for the spectral layer in SpectFormer architecture as shown in figure-\ref{fig_main}. 
 In the first study, we compare various spectral architectures to develop SpectFormer, such as the Fourier Network (FN), Fourier Gating Network (FGN), Fourier Neural Operator (FNO), and Wavelet Gating Network (WGN) as shown in table \ref{Exp_tab_1}.  The Fourier transform network indicates the spectral layer contains just a Fourier transform instead of a multi-headed self-attention network. Similarly, the Fourier gating network uses a Fourier transform and its contribution is controlled by learnable weight parameters, followed by the inverse Fourier transform. We use neural operator techniques for channel mixing and Fourier transform techniques for token mixing similar to FNO paper \cite{guibas2021efficient}. Wavelet gating network uses a wavelet transform followed by learnable weight parameters to control the wavelet decomposition. 
We observe that the Fourier gating network outperforms all other architectures as it uses a gating technique to control the Fourier features. We have done ablation studies on the ImageNet-1K dataset to evaluate the performance of the SpectFormer architecture.  We conduct this ablation study on the small-size networks in stage architecture as mentioned in table-\ref{tab:architecture}.

We also illustrate the performance differences in using the number of spectral layers($\alpha$) in the SpectFormer  architecture as shown in figure-\ref{fig_main}. We select a vanilla transformer architecture similar to DeIT and we replace the number of attention layers with spectral layers. We choose Deit-Small~\cite{touvron2021training} that has 12 layer architecture with a hidden dimension of 384 and a similar architecture in GFNet\cite{rao2021global} is GFNet-XS. We characterize the study using a hyper-parameter $\alpha$. We select the $\alpha$ value zero for the DeiT-S network and a value of twelve for the GFNet-XS transformer. We fine-tune the $\alpha$ value on ImageNet-1K dataset and find that the ideal $\alpha$ value is four. This result is captured in the table \ref{Exp_tab_2}. We started with different $\alpha$ values such as 2, 4, 6, 8, and 10 for validating the DeIT small network where $\alpha_2$ indicates two layers of spectral and ten (12-4) layers of attention in the architecture, while $\alpha_4$ indicates four layers of spectral and eight (12-4) layers of attention network. 

\begin{table}[htbp]
  \centering
  \caption{\textbf{Results on transfer learning datasets}. We report the top-1 accuracy on the four datasets as well as the number of parameters and FLOPs. } 
\setlength{\tabcolsep}{4.0pt}
    \begin{tabular}{c| cccc}
    \toprule
    Model  &   \tabincell{c}{CIFAR\\10}   & \tabincell{c}{CIFAR\\100} & \tabincell{c}{Flowers\\102} & \tabincell{c}{Cars\\196} \\\midrule
    ResNet50~\cite{he2016deep}    & - & - & 96.2  & 90.0 \\
    ViT-B/16~\cite{dosovitskiy2020image}  & 98.1  & 87.1  & 89.5  & - \\
    ViT-L/16~\cite{dosovitskiy2020image}       & 97.9  & 86.4  & 89.7  & - \\
    Deit-B/16~\cite{touvron2021training}     & \textbf{99.1}  & \textbf{90.8}  & 98.4  & 92.1 \\
    ResMLP-24~\cite{touvron2022resmlp}    & 98.7  & 89.5  & 97.9  & 89.5 \\       
     
     GFNet-XS~\cite{rao2021global}   &  98.6  &  89.1  &   98.1  &  92.8 \\
      GFNet-H-B~\cite{rao2021global}   &  99.0  &  90.3  &   98.8  & 93.2 \\\midrule
     Spectformer-B   &  98.9  &  90.3  &   \textbf{98.9}  &  \textbf{93.7} \\
     \bottomrule
    \end{tabular}%
 \vspace{-0.2in}
  \label{tab:transfer_learning}%
\end{table}%

\subsection{Comparison with Similar Architectures}
We compared the vanilla architecture of SpectFormer to the hierarchical architecture of SpectFormer in two parts of the table-\ref{tab:variants}. In the vanilla architecture, we developed tiny (SpectFormer-T), extra small (SpectFormer-XS), small (SpectFormer-S), and base (SpectFormer-B) models that are similar in layer count and hidden dimensions to GFNet~\cite{rao2021global}, while the attention blocks are similar to Deit~\cite{touvron2021training}. Similarly, in the hierarchical architecture, we developed small (SpectFormer-H-S), base (SpectFormer-H-B), and large (SpectFormer-H-L) models using the Fourier gating network. We also developed small and base models using the Fourier and wavelet gating networks, as shown in table-\ref{tab:variants}. We observed that all the hierarchical models (SpectFormer-H-S, SpectFormer-H-B, and SpectFormer-H-L) performed better than the vanilla architecture and are state-of-the-art, as shown in table-\ref{tab:imagenet1k_sota}. 
We compared the performance of SpectFormer to similar architectures on the ImageNet-1k dataset as shown in table-\ref{tab:similar_arch_imagenet}. We first compared SpectFormer to vanilla vision transformers such as DeiT~\cite{touvron2021training}, Fnet~\cite{lee2021fnet}, FourierFormer~\cite{nguyen2022fourierformer}, and GFNet~\cite{rao2021global}, as well as hybrid transformers such as PVT~\cite{wang2021pyramid} and Swin~\cite{liu2021swin} transformers. Compared to attention-based models like DeiT~\cite{touvron2021training}, SpectFormer performed better than DeiT in all size models (T, XS, S, and B). Compared to spectral-based models like Fnet~\cite{lee2021fnet}, FourierFormer~\cite{nguyen2022fourierformer}, and GFNet~\cite{rao2021global}, SpectFormer performed better than all of them in all sizes (T, XS, S, and B). Then, we compared SpectFormer to hierarchical attention architectures such as PVT~\cite{wang2021pyramid}, Swin~\cite{liu2021swin}, LiT~\cite{pan2022less},  and LiTv2\cite{panfast} and spectral architecture GFNet-H-S/B\cite{rao2021global}. We have observed that SpectFormer outperforms vanilla transformers, hybrid transformers, other spectral transformers, and even other weighted attention transformers. SpectFormer performs 2\% better than the latest similar model LiTv1~\cite{pan2022less} for small architecture and 3\% better than the best spectral architecture GFNet-H-S\cite{rao2021global}. When compared to DeiT~\cite{touvron2021training}, SpectFormer also performs better.

\begin{table*}[tb]
\centering
\caption{The performances of various vision models on the COCO val2017 dataset for the downstream tasks of object detection and instance segmentation. RetinaNet is used as the object detector for the object detection task, and the Average Precision ($AP$) at different IoU thresholds or two different object sizes (\emph{i.e.}, small and base) are reported for evaluation. For instance segmentation task, we adopt Mask R-CNN as the base model, and the bounding box and mask Average Precision (\emph{i.e.}, $AP^b$ and $AP^m$) are reported for evaluation.}

\begin{tabular}{c|cccccc|cccccc}
\Xhline{2\arrayrulewidth}
\multirow{2}{*}{Backbone}    & \multicolumn{6}{c}{Mask R-CNN 1x \cite{he2017mask}}     & \multicolumn{6}{c|}{RetinaNet 1x \cite{lin2017focal}}       \\ \cline{2-13}
    & $AP^b$ & $AP^b_{50}$ & $AP^b_{75}$ & $AP^m$  & $AP^m_{50}$ & $AP^m_{75}$ & $AP$ & $AP_{50}$ & $AP_{75}$ & $AP_S$ & $AP_M$ & $AP_L$ \\ \hline
ResNet50 \cite{he2016deep}                  & 38.0 & 58.6  & 41.4  & 34.4 & 55.1  & 36.7  & 36.3 & 55.3 & 38.6 & 19.3 & 40.0 & 48.8  \\
Swin-T   \cite{liu2021swin}                  & 42.2 & 64.6  & 46.2  & 39.1 & 61.6  & 42.0 & 41.5 & 62.1 & 44.2 & 25.1 & 44.9 & 55.5  \\
Twins-SVT-S \cite{chu2021twins}               & 43.4 & 66.0  & 47.3  & 40.3 & 63.2  & 43.4 & 43.0 & 64.2 & 46.3 & 28.0 & 46.4 & 57.5 \\
LITv2-S~\cite{panfast} & 44.9  &-&-&40.8  &-&- &44.0 &-&-  &-&- &-  \\
RegionViT-S \cite{chen2022regionvit}          & 44.2 & -     & -     & 40.8 & -     & -     & 43.9 & -    & -    & -    & -    & -   \\
PVTv2-B2  \cite{wang2022pvt}               
& 45.3 & 67.1  & 49.6  & 41.2 & 64.2  & 44.4  & \textbf{44.6} & \textbf{65.6} & \textbf{47.6} & \textbf{27.4} & \textbf{48.8} & 58.6 \\

SpectFormer-S-FN
& \textbf{46.2}& \textbf{68.1}& \textbf{50.8}& \textbf{42.0} &\textbf{65.2} &\textbf{45.4}  &44.2 & 64.8 & 47.3 & 27.3 & 48.1 & \textbf{59.5}\\

\hline \hline
ResNet101 \cite{he2016deep}                  & 40.4 & 61.1  & 44.2  & 40.4 & 61.1  & 44.2  & 38.5 & 57.8 & 41.2 & 21.4 & 42.6 & 51.1 \\
Swin-S \cite{liu2021swin}                     & 44.8 & 66.6  & 48.9  & 40.9 & 63.4  & 44.2 & 44.5 & 65.7 & 47.5 & 27.4 & 48.0 & 59.9 \\
Twins-SVT-B \cite{chu2021twins}              & 45.2 & 67.6  & 49.3  & 41.5 & 64.5  & 44.8 & 45.3 & 66.7 & 48.1 & 28.5 & 48.9 & 60.6  \\
RegionViT-B \cite{chen2022regionvit}          & 45.4 & -     & -     & 41.6 & -     & -  & 44.6 & -    & -    & -    & -    & -      \\
LITv2-M~\cite{panfast}&46.8 &-&- &42.3&-&-& 46.0&-&- &-&- &- \\
PVTv2-B3  \cite{wang2022pvt}                & 47.0 & 68.1  & 51.7  & 42.5 & 65.7  & 45.7 & 45.9 &\textbf{ 66.8} & 49.3 & 28.6 &\textbf{ 49.8} & \textbf{61.4} \\
SpectFormer-B-FN                            & 46.9 & \textbf{68.8}  & \textbf{51.8}  & \textbf{42.7} & \textbf{65.9}  &\textbf{ 45.7}  &\textbf{ 46.0} & 66.4 & \textbf{49.7} & \textbf{29.5} & 49.7 & 61.1 \\ \Xhline{2\arrayrulewidth}
\end{tabular}
\label{tab:task_learning_1}
\end{table*}


   




\begin{table*}[t]
  \centering
\caption{We conducted a comparison of various transformer-style architectures for image classification on ImageNet. This includes \textbf{vision transformers~\cite{touvron2021training}, MLP-like models~\cite{touvron2022resmlp,liu2021pay}, spectral transformers ~\cite{rao2021global} and our SpectFormer models}, which have similar numbers of parameters and FLOPs. The top-1 accuracy on ImageNet's validation set, as well as the number of parameters and FLOPs, are reported. All models were trained using $224 \times 224$ images. We used the notation "↑384" to indicate models fine-tuned on $384 \times 384$ images for 30 epochs. }\vspace{5pt}
    \begin{tabular}{lrrrrrr}\toprule
    Model & Params (M) & FLOPs (G) & Resolution & Top-1 Acc. (\%) & Top-5  Acc.  (\%) \\ \midrule
    gMLP-Ti~\cite{liu2021pay} & 6     & 1.4   & 224   & 72.0  & - \\
     DeiT-Ti~\cite{touvron2021training} & 5    & 1.2  & 224   & 72.2 &  91.1  \\     
      GFNet-Ti~\cite{rao2021global}  & 7 & 1.3 & 224  & 74.6 & 92.2 \\
     \rowcolor{gray!15} SpectFormer-T & 9 & 1.8 & 224 &  76.8 & 93.3\\ 
     \midrule
     ResMLP-12~\cite{touvron2022resmlp} & 15    & 3.0   & 224   & 76.6  & - \\
      GFNet-XS~\cite{rao2021global} & 16 & 2.9 & 224 & 78.6 & 94.2\\ 
     
\rowcolor{gray!15} SpectFormer-XS & 20& 4.0& 224& 80.2 &94.7\\\midrule
     DeiT-S~\cite{touvron2021training} & 22    & 4.6   & 224   & 79.8  & 95.0  \\
     gMLP-S~\cite{liu2021pay} & 20    & 4.5   & 224   & 79.4  & - \\
     GFNet-S~\cite{rao2021global} & 25 & 4.5 & 224 & 80.0 & 94.9\\ 
     
 \rowcolor{gray!15}SpectFormer-S & 32& 6.6& 224& 81.7 &95.6\\\midrule
     ResMLP-36~\cite{touvron2022resmlp} & 45    & 8.9   & 224   & 79.7  & - \\
     GFNet-B~\cite{rao2021global} & 43 & 7.9 & 224 & 80.7 & 95.1 \\ 
     gMLP-B~\cite{liu2021pay} & 73    & 15.8  & 224   & 81.6  & - \\
    DeiT-B~\cite{touvron2021training} & 86    & 17.5  & 224   & 81.8  & 95.6 \\
     
    \rowcolor{gray!15}  SpectFormer-B &57& 11.5& 224 &\textbf{82.1}& \textbf{95.7}\\ 
     \midrule

    GFNet-XS↑384~\cite{rao2021global} & 18 & 8.4& 384 &  80.6 & 95.4\\ 
    GFNet-S↑384 ~\cite{rao2021global}& 28 &  13.2 & 384 & 81.7 & 95.8\\
    GFNet-B↑384 ~\cite{rao2021global} & 47 & 23.3 & 384 & 82.1 & 95.8\\ 
    \rowcolor{gray!15} SpectFormer-XS↑384 & 21 & 13.1& 384 &  82.1 & 95.7\\ 
     \rowcolor{gray!15} SpectFormer-S↑384 & 33 &  22.0 & 384 & 83.0 & 96.3\\
     \rowcolor{gray!15} SpectFormer-B↑384  & 57 & 37.3 & 384 & 82.9 & 96.1\\ \bottomrule
    \end{tabular}%
  \label{tab:finetune} 
\end{table*}%

\subsection{Image Classification task on ImageNet-1K}
\textbf{Dataset and Training Setups}
We describe the training process of the image recognition task  using the ImageNet1K benchmark dataset, which includes 1.28 million training images and 50K validation images belonging to 1,000 categories. The vision backbones are trained from scratch using data augmentation techniques like RandAug, CutOut, and Token Labeling objectives with MixToken. The performance of the trained backbones is evaluated using both top-1 and top-5 accuracies on the validation set. The optimization process involves using the AdamW optimizer with a momentum of 0.9, 10 epochs of linear warm-up, and 310 epochs of cosine decay learning rate scheduler. The batch size is set to 128 and is distributed on 8 A100 GPUs. The learning rate and weight decay are fixed at 0.00001 and 0.05, respectively.

Table-\ref{tab:imagenet1k_sota} presents a comparison of the performance of the state-of-the-art vision models and our SpectFormer variants. The ViT backbones with the best performance, VOLO-D1$^\star$, VOLO-D2$^\star$, and VOLO-D3$^\star$, are trained using additional strategies such as Token Labeling objective with MixToken and convolutional stem for better patch encoding. We also use these strategies to train our SpectFormer variants in each size, which are denoted as SpectFormer-S$^\star$, SpectFormer-B$^\star$, and SpectFormer-L$^\star$. For a fair comparison, we degraded the version of SpectFormer in Small size without token labeling objective and which is called SpectFormer-S and its top-1 accuracy is 83.1 which is better than Wave-ViT-s (82.7, without extra token).

The table shows that our Wave-ViT variants consistently outperform existing vision models, including ResNet, SE-ResNet, Vanilla ViTs (TNT, CaiT, CrossViT), and hierarchical ViTs (Swin, Twins-SVT, PVTv2, VOLO), under similar GFLOPs for each group.
 In particular, under the Base size, the Top-1 accuracy score of SpectFormer-B$^\star$ can reach 85.1\%, which leads to the absolute improvement of 0.3\% against the best competitive Wave-ViT-B$^\star$ (Top-1 accuracy: 84.8\%). Under the Large size, when compared to ResNet-152 and SE-ResNet-152, which solely rely on CNN architectures, vanilla ViTs (TNT-B, CaiT-S36, and CrossViT) capture long-range dependencies through Transformer structure and outperform them. However, the performances of CaiT-S36 and CrossViT are still lower than most hierarchical ViTs (PVTv2-B5, VOLO-D3$^\star$, CMT-L, MaxViT, DaViT and Wave-ViT-L ) that aggregate multi-scale contexts. Moreover, unlike PVTv2-B5, which uses irreversible down-sampling for self-attention learning, Wave-ViT uses invertible down-sampling with wavelet transforms for self-attention learning.  In particular, under the large size, the Top-1 accuracy score of SpectFormer-L$^\star$ can reach 85.7\%, which leads to the absolute improvement of 0.2\% against the best competitive Wave-ViT-L$^\star$ (Top-1 accuracy: 85.5\%). Our SpectFormer-L$^\star$ achieves better efficiency-vs-accuracy trade-off by enabling initial spectral blocks in  the transformer encoder and attention blocks are at the top blocks. Overall, these findings demonstrate the efficacy of spectral block along with attention block in enhancing visual representation learning.

\begin{table*}[]
  \centering
  \caption{This table presents information about datasets used for transfer learning. It includes the size of the training and test sets, as well as the number of categories included in each dataset.  }
\setlength{\tabcolsep}{4.0pt}
    \begin{tabular}{c| c|c|c|c}
    \toprule
    Dataset  &   CIFAR-10~\cite{krizhevsky2009learning}   & CIFAR-100~\cite{krizhevsky2009learning} & Flowers-102~\cite{nilsback2008automated} & Stanford Cars~\cite{krause20133d} \\\midrule
    Train Size  & 50,000  & 50,000 & 8,144 & 2,040\\
    Test Size  & 10,000  & 10,000 & 8,041  & 6,149\\
    \#Categories & 10& 100 & 196 & 102\\
     \bottomrule
    \end{tabular}%
 \vspace{-1.12em}
  \label{tab:transfer_learning_dataset}%
\end{table*}%
\begin{table}[!tb]
\centering
\caption{The performances of various vision backbones on COCO val2017 dataset for the downstream task of object detection. Four kinds of object detectors, \emph{i.e.}, GFL ~\cite{li2020generalized}, and  Cascade Mask R-CNN~\cite{cai2018cascade} in mmdetection \cite{chen2019mmdetection}, are adopted for evaluation. We report the bounding box Average Precision ($AP^b$) in different IoU thresholds.}
\begin{tabular}{c|c|ccc}
\Xhline{2\arrayrulewidth}
Backbone   & Method & $AP^b$ & $AP^b_{50}$ & $AP^b_{75}$  \\ \hline

ResNet50 \cite{he2016deep}  & \multirow{4}{*}{GFL \cite{li2020generalized}}
   & 44.5 & 63.0  & 48.3  \\
   
Swin-T   \cite{liu2021swin}     &    & 47.6 & 66.8  & 51.7 \\
PVTv2-B2  \cite{wang2022pvt}  &    & 50.2 & 69.4  & 54.7 \\
SpectFormer-H-S-FN                 &  & \textbf{50.3} & \textbf{70.0} & \textbf{55.2}\\ 

\hline\hline

ResNet50 \cite{he2016deep}  & \multirow{4}{*}{\begin{tabular}[c]{@{}c@{}}Cascade\\ Mask\\ R-CNN\end{tabular} \cite{cai2018cascade}} & 46.3 & 64.3  & 50.5  \\
Swin-T \cite{liu2021swin}    &       & 50.5 & 69.3  & 54.9  \\
PVTv2-B2 \cite{wang2022pvt}  &     & 51.1 & 69.8  & 55.3   \\

SpectFormer-H-S-FN &                   &\textbf{ 51.5} & \textbf{70.2}  & \textbf{56.3 }  \\ \Xhline{2\arrayrulewidth}
\end{tabular}
\label{tab:task_learning_2}
\vspace{-0.9em}
\end{table}
\subsection{Task Learning: Object Detection}

\textbf{Training setup: }
In this section, we examine the pre-trained SpectFormer-H-small behavior on COCO dataset for two downstream tasks that localize objects ranging from bounding-box level to pixel level, \emph{i.e.}, object detection and instance segmentation. Two mainstream detectors,  \emph{i.e.}, RetinaNet \cite{lin2017focal} and Mask R-CNN\cite{he2017mask} as shown in table-8 of the main paper, and two state-of-the-art detectors \emph{i.e.},  GFL ~\cite{li2020generalized}, and  Cascade Mask R-CNN~\cite{cai2018cascade} in mmdetection \cite{chen2019mmdetection} in this supplementary doc.
We are employed for each downstream task, and we replace the CNN backbones in each detector with our SpectFormer-H-small for evaluation. Specifically, each vision backbone is first pre-trained over ImageNet1K, and the newly added layers are initialized with Xavier \cite{glorot2010understanding}. Next, we follow the standard setups in \cite{liu2021swin} to train all models on the COCO train2017 ($\sim$118K images). Here the batch size is set as 16, and AdamW \cite{loshchilovdecoupled} is utilized for optimization (weight decay: 0.05, initial learning rate: 0.0001, betas=(0.9, 0.999)).  We used learning rate (lr) configuration with step lr policy, linear warmup at every 500 iterations with warmup ration 0.001. All models are finally evaluated on the COCO val2017 (5K images).  For state-of-the-art models like GFL ~\cite{li2020generalized}, and  Cascade Mask R-CNN~\cite{cai2018cascade}, we utilize 3 $\times$ schedule (\emph{i.e.}, 36 epochs) with the multi-scale strategy for training, whereas for RetinaNet \cite{lin2017focal} and Mask R-CNN\cite{he2017mask} we utilize 1 $\times$ schedule (\emph{i.e.}, 12 epochs).

We conducted experiments on MS COCO 2017, which is a widely used benchmark for object detection and instance segmentation, comprising around 118K images for the training set and approximately 5K images for the validation set. Our approach involved experimenting with two detection frameworks, namely RetinaNet \cite{lin2017focal} and Mask R-CNN\cite{he2017mask}, and we measured model performance using Average Precision (AP). 
We use the pre-trained model SpectFormer trained on the ImageNet-1K dataset to initialize the backbone architecture and Xavier initialization for additional layers of the network. These results are shown in table-~\ref{tab:task_learning_1}. The experimental results, as presented in table-~\ref{tab:task_learning_1}, indicate that SpectFormer has comparative results on both the RetinaNet \cite{lin2017focal} and Mask R-CNN\cite{he2017mask} models. We have compared with the latest work including LITv2~\cite{panfast}, RegionViT~\cite{chen2022regionvit}, and PVT~\cite{wang2021pyramid} transformer models. Further, our SpectFormer model demonstrated significantly better performance than ResNet in terms of AP. More importantly, SpectFormer outperformed all compared vanilla ViT models and hierarchical transformer models, achieving the best AP performance.  We compared the performance of SpectFormer with other hierarchical models such as Swin-T \cite{liu2022swin} and Pyramid Vision Transformer (PVT) \cite{wang2022pvt} for two state-of-the-art object detection tasks. The results are presented in Table \ref{tab:task_learning_2}, and demonstrate that SpectFormer outperforms other SOTA transformer-based models for these object detection models. Additionally, we report the object detection performance of GFL\cite{li2020generalized} model and Cascade Mask R-CNN~\cite{cai2018cascade} on MS COCO val2017 dataset, which demonstrates an improvement in performance, as shown in Table-\ref{tab:task_learning_2}.

\begin{figure*}[htb]%
\centering
\includegraphics[width=0.949\textwidth]{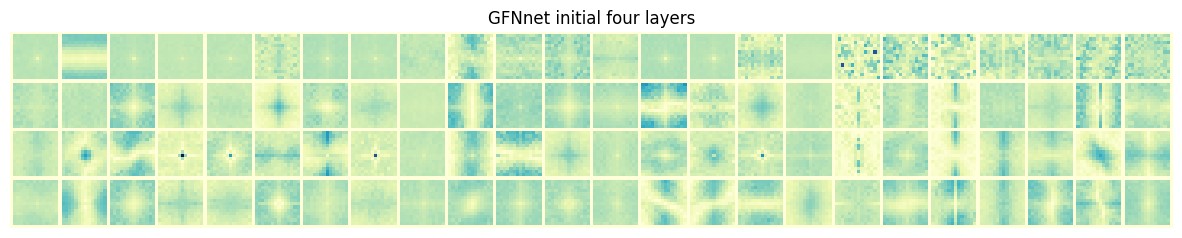}\\
\includegraphics[width=0.949\textwidth]{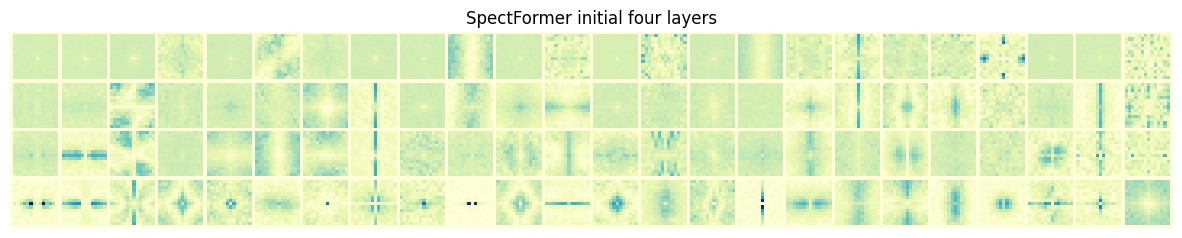}
\caption{This figure shows the Filter characterization of the initial four layers of  the GFNet~\cite{rao2021global} and SpectFormer model. It clearly shows that spectFormer captures local filter information such as lines and edges of an Image.}\label{fig_filter}
\end{figure*}

\begin{figure*}[htb]%
\centering
\includegraphics[width=0.989\textwidth]{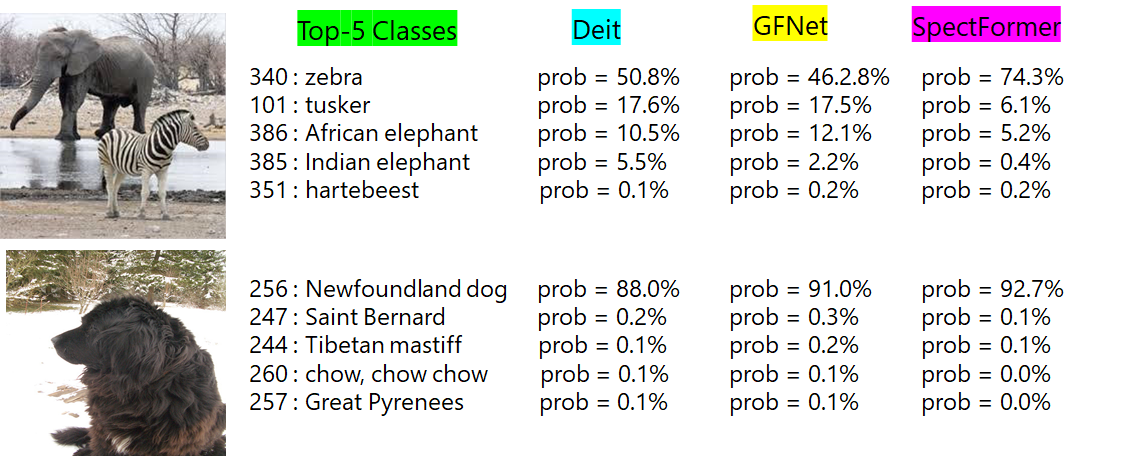}\\
\caption{The figure shows the top-5 class prediction probability scores for Deit \cite{touvron2021training}, GFNet \cite{rao2021global}, and our SpectFormer model, indicating that SpectFormer predicts the 'Zebra' class (Top-1 Class) with greater confidence than GFNet and Deit.}\label{fig_prob_score}
\vspace{-1.0em}
\end{figure*}

 \subsection{Transfer Learning Comparison}
 \textbf{Training setup:}
To test the effectiveness of our architecture and learned representation, we evaluated vanilla SpectFormer on commonly used transfer learning benchmark datasets, including CIFAR-10~\cite{krizhevsky2009learning}, CIFAR100~\cite{krizhevsky2009learning}, Oxford-IIIT-Flower~\cite{nilsback2008automated} and Standford Cars~\cite{krause20133d}. Our approach followed the methodology of previous studies~\cite{tan2019efficientnet,dosovitskiy2020image,touvron2021training,touvron2022resmlp,rao2021global}, where we initialized the model with ImageNet pre-trained weights and fine-tuned it on the new datasets. In table-7 of the main paper, we have presented a comparison of the transfer learning performance of our basic and best models with state-of-the-art CNNs and vision transformers. The transfer learning setup employs a batch size of 64, a learning rate (lr) of 0.0001, a weight-decay of 1e-4, a clip-grad of 1, and warmup epochs of 5. We have utilized a pre-trained model trained on the Imagenet-1K dataset, which we have fine-tuned on the transfer learning dataset specified in table-\ref{tab:transfer_learning} for 1000 epochs.  Table-\ref{tab:transfer_learning_dataset} displays the dataset information used for transformer learning.

In order to assess the effectiveness of SpectFormer's architecture and learned representation, we conducted evaluations on multiple transfer learning benchmark datasets, which included CIFAR-10~\cite{krizhevsky2009learning}, CIFAR-100~\cite{krizhevsky2009learning}, Stanford Cars~\cite{krause20133d}, and Flowers-102~\cite{nilsback2008automated}. 
Here we compare the performance of SpectFormer pre-trained on ImageNet-1K and fine-tuned on the new datasets for the image classification task. Both the basic and best models were evaluated for their transfer learning performance, and  the comparison is captured in table-~\ref{tab:transfer_learning}. The results show that the proposed models performed well on downstream datasets, surpassing ResMLP models by a significant margin and achieving highly competitive performance comparable to state-of-the-art spectral network, GFNet\cite{rao2021global}. Our models also exhibited competitive performance when compared to state-of-the-art CNNs and vision transformers.



\subsection{Model Fine-tuning for High Resolution input}
Our main experiments are conducted on ImageNet~\cite{deng2009imagenet}, a popular benchmark for large-scale image classification. To ensure a fair comparison with previous research~\cite{touvron2021training,touvron2022resmlp,rao2021global}, we adopt the same training details for our SpectFormer models. For the vanilla transformer architecture (SpectFormer), we use the hyper-parameters recommended by the GFNet implementation~\cite{rao2021global}. For the hierarchical architecture (SpectFormer-H), we use the hyper-parameters recommended by the WaveVit implementation~\cite{yao2022wave}. We use the hyper-parameters recommended by the GFNet implementation~\cite{rao2021global} and train our models for 30 epochs during fine-tuning at higher resolutions. All models are trained on a single machine equipped with 8 A100 GPUs. In our experiments, we compared the fine-tuning performance of our models with GFNet~\cite{rao2021global}. Our observations indicate that our SpectFormer model outperforms GFNet's base spectral network. Specifically, SpectFormer-S(384) achieves a performance of 83.0\%, which is 1.2\% higher than GFNet-S(384), as shown in Table~\ref{tab:finetune}. Similarly, SpectFormer-XS and SpectFormer-B perform better than GFNet-XS and GFNet-B, respectively. In the results section, we present the fine-tuned results for models trained on 224 x 224 and fine-tuned on 384 x 384, as depicted in Table-\ref{tab:finetune}.




\subsection{Visualization of filter weights of spectral layers}

Filter characterization is a technique to analyze the learned filters in transformer networks and gain insight into what kind of features the models is learning at different layers. By visualizing the learned filters, we can get a better understanding of how the transformer is processing images.

In our experiments, we performed filter characterization visualization for the initial four layers of GFNet~\cite{rao2021global} and SpectFormer as shown in figure-\ref{fig_filter}.
In this figure, we display the initial 24 filters of each layer and compare them between these two models. We observed that SpectFormer captures local filter information, such as lines and edges of an image, more clearly than GFNet. This suggests that SpectFormer is better able to capture local image details and, therefore, may be better suited for tasks that require high-resolution image processing.

\section{Conclusion}
Through this work, we analysed the core architecture of transformers by using a mixed approach that includes spectral and multi-headed attention. Previously, transformers used either all-attention layers or more recently spectral layers have been used. Spectformer combines both these aspects and shows consistently better performance than either all-attention or all-spectral layers. We use a parameterized approach that suggests further scope for adaptation of this work for specific tasks. For instance, work in remote-sensing or medical-imaging may choose a different combination of spectral and attention layers to obtain best performance. 
The work achieves state-of-the-art (85.7\%) top-1 accuracy on ImageNet-1K dataset. 

{\small
\bibliographystyle{ieee_fullname}
\bibliography{egbib}
}

\end{document}